# INSIGHT INTO THE COLLOCATION OF MULTI-SOURCE SATELLITE IMAGERY FOR MULTI-SCALE VESSEL DETECTION

*Tran-Vu La[1], Minh-Tan Pham[2], and Marco Chini[1]*

[1]Luxembourg Institute of Science and Technology, Esch-sur-Alzette, Luxembourg
[2]IRISA, Université Bretagne Sud, UMR 6074, 56000 Vannes, France

## ABSTRACT

Ship detection from satellite imagery using Deep Learning (DL) is an indispensable solution for maritime surveillance. However, applying DL models trained on one dataset to others having differences in spatial resolution and radiometric features requires many adjustments. To overcome this issue, this paper focused on the DL models trained on datasets that consist of different optical images and a combination of radar and optical data. When dealing with a limited number of training images, the performance of DL models via this approach was satisfactory. They could improve 5–20% of average precision, depending on the optical images tested. Likewise, DL models trained on the combined optical and radar dataset could be applied to both optical and radar images. Our experiments showed that the models trained on an optical dataset could be used for radar images, while those trained on a radar dataset offered very poor scores when applied to optical images.

## 1. INTRODUCTION

Ship detection from satellite imagery has an indispensable role in maritime surveillance, thanks to its large coverage, high resolution, availability, and accessibility. Furthermore, Synthetic Aperture Radar (SAR) satellite sensors can acquire data in most weather conditions, while optical devices can detect quite accurately the size and shape of vessels. Nevertheless, one of the main constraints of using optical images for ship detection is cloud impact which can lead to the lack of important information. Additionally, the small-scale cloud patterns having similar size and pixel values to the vessels can produce misclassifications. Regarding SAR images, ship detection can be affected by speckle noise, azimuth ambiguity, and convective cells. Likewise, strong radar reflectance from the ships can make difficulties in the observation of their shape and size. One of the practical and effective solutions to overcome these issues is to combine multi-source satellite imagery. This approach allows not only to benefit the advantages of different sensors but also to improve the availability of data for ship detection, especially for emergency cases.

Ship detection from optical imagery has been widely investigated in the last ten years, especially based on Deep Learning (DL) models. However, most studies on that topic focused on very high-resolution (VHR) optical imagery [1-4] having pixel spacing below 1 m, while few [5-7] proposed the detection of vessels from high (HR) or moderate (MR) spatial resolution, i.e., 2–30 m pixel spacing. It should be noted that most MR data sources are freely accessible, while VHR images are commercial and hard to access. Detecting ships with dimensions below the image pixel size is generally not feasible. However, in most practical cases, most vessels have a length and width superior to the pixel sizes of satellite imagery. Therefore, HR and MR optical images can be exploited to detect multi-scale ships. Regarding SAR-based ship detection, the studies can be divided into two directions, including adaptive thresholding algorithms, e.g., Constant False Alarm Rate (CFAR) [8], and DL models [3-4]. The CFAR approach can be easily implemented for different SAR datasets; however, the accuracy of ship detection cannot be satisfactory in complex cases since it does not include many parameters to adjust [8].

The DL models can detect vessels more accurately and generally, but they require time and effort to implement, especially for the preparation of training, validation, and test sets. This issue is more significant if we use different satellite imagery datasets for ship detection because data preparation tasks will be multiplied. Therefore, this paper aims to investigate various approaches to apply the DL models trained and tested on the datasets having differences in spatial resolution and radiometric features. Concretely, we focus on two aspects of ship detection from multi-source satellite imagery that have not been attentively discussed in the literature. First, we compare the performance of DL models trained on one HR or MR dataset and those trained on the combined HR and MR datasets. Second, we compare the performance of DL models trained on an optical or SAR dataset and tested on another one. Likewise, we evaluate the performance of DL models trained on the combined SAR and optical dataset. The objective of this work is to answer a practical question of ship detection in maritime surveillance, especially for emergency cases, if we can directly apply the

DL models trained on one dataset to others having differences in spatial resolution and radiometric features, without the supplementary steps such as data preparation and DL models retraining.

## 2. METHODOLOGY

### 2.1. Data selection and preparation

The HR and MR optical data used for ship detection in this study include PlanetScope, Sentinel-2, Gaofen-1/6, and Landsat-8. Their spatial resolution varies from 2 m (PlanetScope) to 30 m (Landsat). Sentinel-2 and Gaofen-1/6 have close spatial resolutions, i.e., 10 m vs. 16 m. For SAR, we used Sentinel-1 images having the same spatial resolution as Sentinel-2, i.e., 10 m pixel spacing. As most DL models for ship detection are trained on the Red-Green-Blue (RGB) images (8 bits for each channel), we need to transform the original data to this data type. For optical images, the three channels RGB are derived from the reflectance values of Red, Green, and Blue bands which are scaled to the range of (0–0.3), while the RGB images of Sentinel-1 SAR data are obtained from the radar backscattering or Normalized Radar Cross Section (NRCS) of vertical co-polarization (VV), vertical–horizontal cross-polarization (VH), and ratio of VV/VH, respectively.

For training DL models, the common approach is to divide the whole image into sub-images to reduce processing time. We selected the sub-image size of 512 × 512 pixels for this study. The training, validation, and test set for one optical or SAR dataset consists of 200, 67, and 67 sub-images, respectively. Accordingly, the four-combined optical dataset includes 800, 268, and 268 sub-images, and the SAR-optical combined dataset consists of 400, 134, and 134 sub-images. We first test our ideas on a small dataset in this paper and then generalize it on a large one in future studies.

### 2.2. Deep learning models for ship detection

Ship detection from satellite imagery using DL models has been widely investigated in recent years [1-4]. Among them, YOLO (You Only Look Once) with different versions is one of the most used DL models for ship detection [1]. It is a comprehensive DL detection model that swiftly identifies object-bound boxes and classifies them in a single step. In YOLO, the input image is initially divided into a grid of non-overlapping cells, from which three key elements are predicted: (1) the likelihood of an object's presence; (2) the coordinates of its bounding box if an object exists; and (3) the object class and confidence score. In this paper, we will test our ideas with YOLOv4 [9] and YOLOv5s [10], as well as with the DRENet model [6] which has been recently proposed based on YOLO structure in the context of ship detection from Gaofen images.

To evaluate the performance of YOLO models for ship detection, we focus on mean average precision (mAP). The AP is the average precision value computed when the Recall value varies from 0 and 1. When dealing with multiple object classes, the mAP is computed as the mean of AP over all the classes. In the case of detecting a single class such as ship detection in this study, mAP and AP have the same value. More details of AP and mAP calculations can be found in [3].

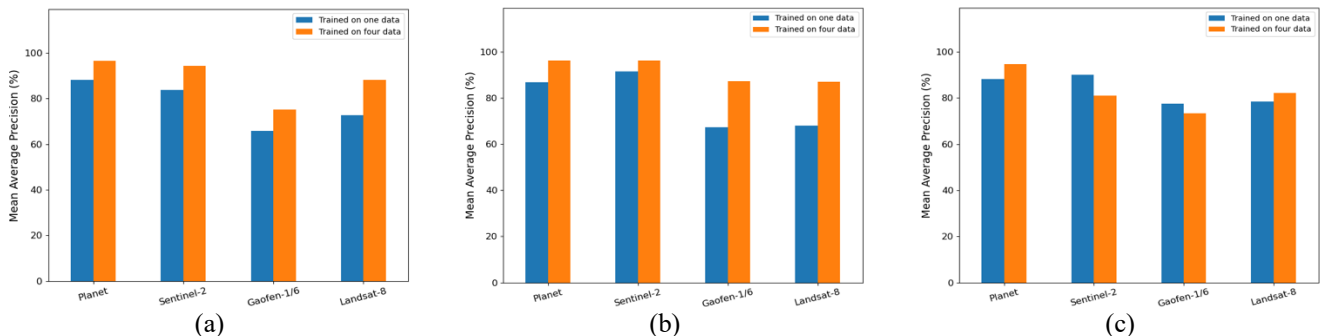

**Fig. 1.** Comparison of the performance (mAP) of DL models trained on one dataset (Planet, Sentinel-2, Gaofen-1/6, Landsat-8) and four-combined datasets (Planet+Sentinel-2+Gaofen-1/6+Landsat-8). (a) YOLOv5s. (b) DRENet. (c) YOLOv4.

## 3. EXPERIMENTAL RESULTS

### 3.1. DL models trained one optical dataset vs. the four-combined dataset

We compare the performance of YOLOv4, YOLOv5s, and DRENet trained on one of four optical datasets (Planet, Sentinel-2, Gaofen-1/6, and Landsat-8) to those trained on the four-combined dataset to evaluate the changes in the performance of DL models trained on a dataset consisting of different optical images. As shown in Fig. 1, the mAP of the DL models trained on the four-combined dataset is significantly improved, compared to those trained on one dataset. For Planet and Sentinel-2, the scores increase from 5% to 10% and about 10–20% for Gaofen-1/6 and Landsat-8, depending on the applied models. The performance of YOLOv5s and DRENet trained on the four-combined dataset is better than that of YOLOv4. The results in Fig. 1 showed that the combination of four datasets with a difference in spatial resolution (HR, MR, and LR) enables a significant improvement in ship detection, especially from Gaofen-1/6 and Landsat-8, for which the models trained on only one dataset did not work well (about 65–72%). In other words, the increase in the number of training data, despite other types, can enable a significant improvement in the performance of DL models. This behavior can be explained by the fact that, regardless of the sensors, the appearance of ships on the sea surface is quite similar. Combining multi-sensor images to train deep learning models becomes essential for ship detection to deal with the lack of annotations and the scarcity of some specific high-resolution images for training.

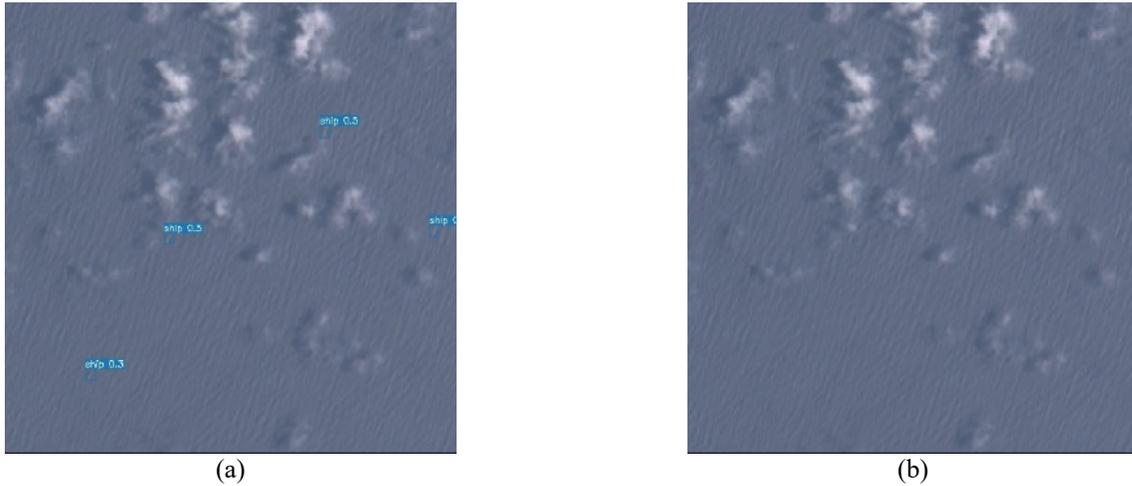

**Fig. 2.** False positives associated with the small-scale cloud patterns on the Sentinel-2 image, June 30, 2022, 06:56:41 UTC, given by (a) DRENet trained on only one Sentinel-2 dataset, compared to (b) DRENet trained on the four-combined dataset. This scene does not consist of any ground truth of ship detection.

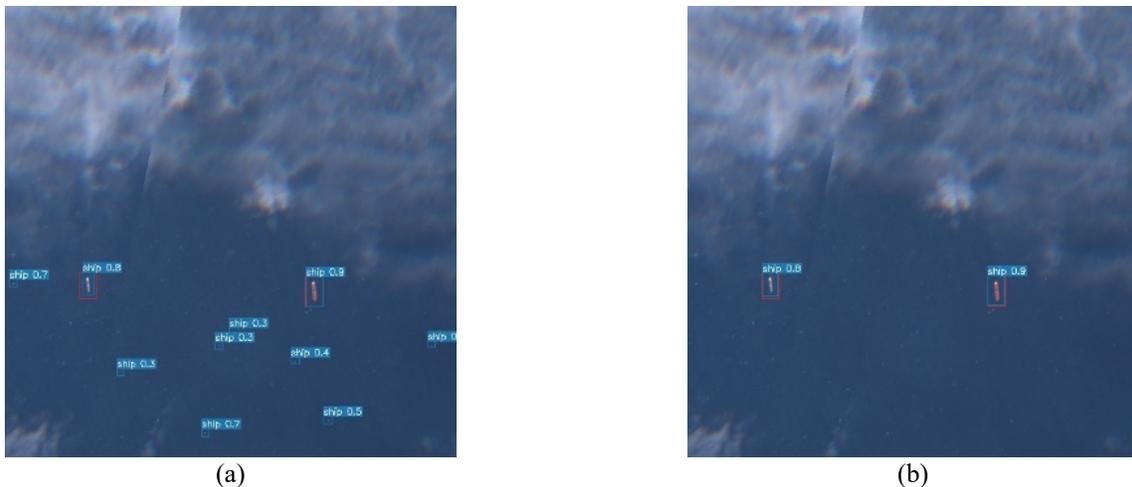

**Fig. 3.** False positives due to the noise (bright pixels) corresponding to significant white foam crests caused by strong surface winds (above 10 m/s) on the Sentinel-2 image, Jan. 16, 2022, 07:02:29 UTC, given by (a) DRENet trained on only one Sentinel-2 dataset, compared to (b) DRENet trained on the four-combined dataset. The red bounding boxes are the ground truth of ship detection.

To understand more deeply the performance improvement of DL models trained the four-combined dataset, Fig. 2 compares the false positives associated with the small-scale cloud patterns on the Sentinel-2 image, June 30, 2022, 06:56:41 UTC, given by DRENet trained on only one Sentinel-2 dataset (Fig. 2a) and the four-combined dataset (Fig. 2b). The latter succeeds to remove all false positives associated with the small-scale cloud patterns, while DRENet trained on only one Sentinel-2 dataset gives four false positives. Likewise, Fig. 3 compares the false positives due to the noise (bright pixels) corresponding to significant white foam crests caused by strong surface winds (above 10 m/s) on the Sentinel-2 image, Jan. 16, 2022, 07:02:29 UTC, given by DRENet trained on only one Sentinel-2 dataset

(Fig. 3a) and the four-combined dataset (Fig. 3b). We realize that while the model trained on only one dataset gives many false positives (eight), the one trained on the four-combined dataset can remove all false positives associated with the white foam crests.

TABLE I
PERFORMANCE (mAP) OF DL MODELS TRAINED ON SENTINEL-1/2 DATASET AND THE COMBINED SENTINEL-1/2 DATASET

| Model | Dataset | Sentinel-1 | Sentinel-2 | S-1+S-2 |
|---|---|---|---|---|
| YOLOv5s | Sentinel-1 | **93.71** | *2.50* | N/A |
| | Sentinel-2 | 85.90 | 83.63 | N/A |
| | S-1+S-2 | 92.40 | **93.77** | **93.88** |
| DRENet | Sentinel-1 | 93.40 | *5.30* | N/A |
| | Sentinel-2 | 89.45 | **91.62** | N/A |
| | S-1+S-2 | **93.86** | 90.30 | **92.11** |

### 3.2. DL models trained on a SAR / optical dataset vs. the combined SAR–optical dataset

This section aims to evaluate the performance of DL models trained on a SAR / optical dataset and tested on others, as well as that trained on the combined SAR and optical dataset. In general, SAR and optical sensors operate with different radiometric techniques, and therefore they observe vessels in different ways. Nevertheless, as the YOLO models work on the RGB images, we expect that it is possible to test the DL models trained on a SAR or optical dataset on others.

Table I presents the performance (mAP) of YOLOv5s and DRENet trained on the Sentinel-1 SAR or Sentinel-2 optical dataset and the combined Sentinel-1/2 dataset. When we apply the models trained on Sentinel-1 to Sentinel-2, their mAP scores are significantly reduced to below 5%. However, when we apply the models trained on Sentinel-2 to Sentinel-1, the scores are still satisfactory, compared to the performance of the models trained on Sentinel-1. This result shows that the application of DL models trained on optical datasets to SAR ones is possible; however, it is not possible if we directly apply the models trained on SAR datasets to optical ones. These behaviors could be explained as follows. Models trained on optical images might perform reasonably well on SAR images because SAR data generally carries more fundamental structural information that might be easier to interpret even without specific training. On the other hand, models trained on SAR data might struggle with optical images because they lack color information, fine details, and visual characteristics significantly present in optical data. Table I also shows that the models trained on the combined Sentinel-1/2 dataset can be applied to Sentinel-1 or Sentinel-2 as their mAP scores are similar to those trained on Sentinel-1 or Sentinel-2.

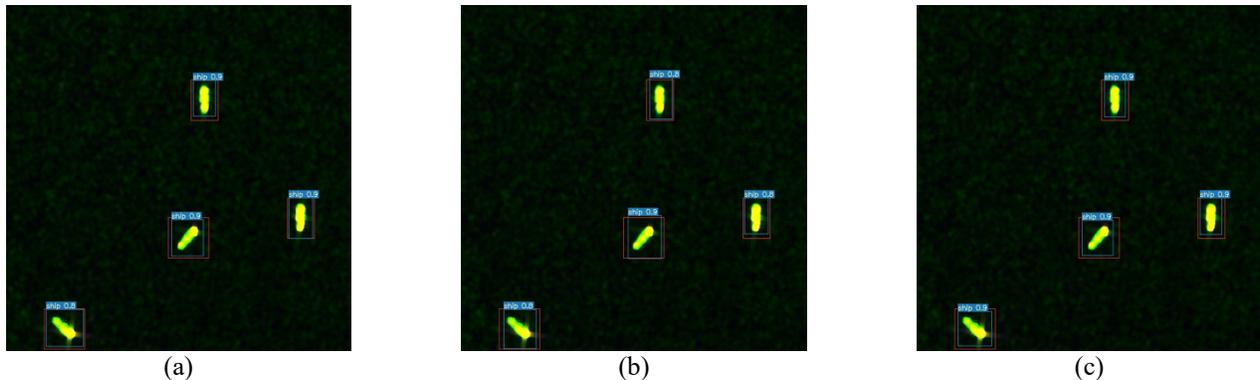

(a)　　　　　　　　　(b)　　　　　　　　　(c)

**Fig. 4.** Ship detection on the Sentinel-1 image, Jan. 6, 2021, 02:23:22, off the coast of Qatar, using DRENet trained on (a) Sentinel-1 SAR dataset, (b) Sentinel-2 optical dataset, and (c) the combined Sentinel-1/2 dataset.

### 4. CONCLUSION

The results showed that the combination of optical images with a difference in spatial resolution (HR, MR, and LR) enabled a significant improvement in ship detection when we applied them to different optical datasets. Additionally, when we applied the DL models trained on an optical (Sentinel-2) dataset to a SAR (Sentinel-1) one, they could offer very good mAP scores. However, the DL models trained on a SAR dataset could not be used for an optical dataset. To solve this issue, we could apply the DL models trained on the combined optical-SAR dataset to SAR ones. Other future works could investigate more advanced DL approaches to foster ship detection performance by exploring models adapted to small objects [11-14] as well as combining multimodal optical-SAR data [15-16].

### 5. ACKNOWLEDGMENT

This work is supported by the Luxembourg National Research Fund (FNR) in the framework of the CORE project C20/SR114703579.

# 6. REFERENCES

[1] M. J. Er, Y. Zhang, J. Chen, and W. Gao, "Ship detection with deep learning: A survey." Artificial Intelligence Review 56, no. 10 (2023): 11825-11865.

[2] L. Li et al., "A Novel CNN-Based Method for Accurate Ship Detection in HR Optical Remote Sensing Images via Rotated Bounding Box," *IEEE Trans. on Geo. & Rem. Sens.*, vol. 59, no. 1, pp. 686-699, Jan. 2021.

[3] Z. Hong et al., "Multi-Scale Ship Detection From SAR and Optical Imagery Via A More Accurate YOLOv3," *IEEE JSTARS*, vol. 14, pp. 6083-6101, 2021.

[4] M. Yasir et al., "Ship detection based on deep learning using SAR imagery: a systematic literature review." Soft Computing 27, no. 1 (2023): 63-84.

[5] Z. Liu et al., "Moving Ship Optimal Association for Maritime Surveillance: Fusing AIS and Sentinel-2 Data," *IEEE Trans. on Geo. & Rem. Sens.*, vol. 60, pp. 1-18, 2022.

[6] J. Chen et al., "A Degraded Reconstruction Enhancement-Based Method for Tiny Ship Detection in Remote Sensing Images With a New Large-Scale Dataset," *IEEE Trans. on Geo. & Rem. Sens.*, vol. 60, pp. 1-14, 2022.

[7] T. V. La, M. T. Pham, and M. Chini, "Collocation of multi-source satellite imagery for ship detection based on Deep Learning models" (No. EGU24-3954). Copernicus Meetings, 2024.

[8] D. J. Crisp, "The State-of-the-Art in Ship Detection in Synthetic Aperture Radar Imagery." DSTO–RR–0272, 2004-05.

[9] A. Bochkovskiy, C. Y. Wang, and H. Y. M. Liao, "Yolov4: Optimal speed and accuracy of object detection," arXiv preprint arXiv:2004.10934, 2020.

[10] G. Jocher et al., Ultralytics YOLOv5: https://docs.ultralytics.com/yolov5. Accessed March 2024.

[11] M. T. Pham, L. Courtrai, C. Friguet, S. Lefèvre, and A. Baussard, "YOLO-Fine: One-stage detector of small objects under various backgrounds in remote sensing images." Remote Sensing 12, no. 15 (2020): 2501.

[12] S. Gui, S. Song, R. Qin and Y. Tang, "Remote Sensing Object Detection in the Deep Learning Era—A Review." Remote Sensing 16, no. 2 (2024): 327.

[13] L. Courtrai, M. T. Pham and S. Lefèvre. "Small object detection in remote sensing images based on super-resolution with auxiliary generative adversarial networks." Remote Sensing 12.19 (2020): 3152.

[14] A. Froidevaux et al., "Vehicle detection and counting from VHR satellite images: efforts and open issues." IGARSS 2020-2020 IEEE International Geoscience and Remote Sensing Symposium. IEEE, 2020.

[15] A. Belmouhcine et al. "Multimodal Object Detection in Remote Sensing." IGARSS 2023-2023 IEEE International Geoscience and Remote Sensing Symposium. IEEE, 2023.

[16] P. Berg et al. "Multimodal Supervised Contrastive Learning in Remote Sensing Downstream Tasks." IEEE Geoscience and Remote Sensing Letters (2024).